\documentclass{article}

\PassOptionsToPackage{numbers, compress}{natbib}
\usepackage[final]{neurips_2025}
\usepackage{natbib}
\PassOptionsToPackage{numbers}{natbib}

\usepackage[table]{xcolor}
\usepackage{multirow}
\usepackage{amsmath}
\usepackage{booktabs} % 提供专业的三线表格式
\usepackage{tabularx} % 用于调整表格宽度
\usepackage{adjustbox}
\usepackage{newtxmath}
\usepackage{arydshln}
\usepackage{colortbl}

\definecolor{RowColor}{rgb}{0.97, 0.97, 1}
\usepackage[utf8]{inputenc} % allow utf-8 input
\usepackage[T1]{fontenc}    % use 8-bit T1 fonts
\usepackage{hyperref}       % hyperlinks
\usepackage{url}            % simple URL typesetting
\usepackage{booktabs}       % professional-quality tables
\usepackage{amsfonts}       % blackboard math symbols
\usepackage{nicefrac}       % compact symbols for 1/2, etc.
\usepackage{microtype}      % microtypography
\usepackage{xcolor}         % colors
\definecolor{lblue}{HTML}{E2F0FF}

\title{Consistent Video Editing as Flow-Driven Image-to-Video Generation}

\author{
  Ge Wang\textsuperscript{1}\thanks{Equal contribution.}\quad
  Songlin Fan\textsuperscript{1}\footnotemark[1]\quad
  Hangxu Liu\textsuperscript{1}\quad
  Quanjian Song\textsuperscript{2}\quad
  Hewei Wang\textsuperscript{3}\quad
  Jinfeng Xu\textsuperscript{4}\thanks{Corresponding author: jinfeng@connect.hku.hk}
  \\
  \textsuperscript{1}Fudan University \hspace{0.1em}
  \textsuperscript{2}Xiamen University \hspace{0.1em}
  \textsuperscript{3}Carnegie Mellon University \hspace{0.1em}
  \textsuperscript{4}The University of Hong Kong
}

\makeatletter
\renewcommand{\@notice}{}
\makeatother

\begin{document}

\maketitle
\vspace{-1.5em}
\begin{figure}[h]
    \centering
    \includegraphics[width=\textwidth]{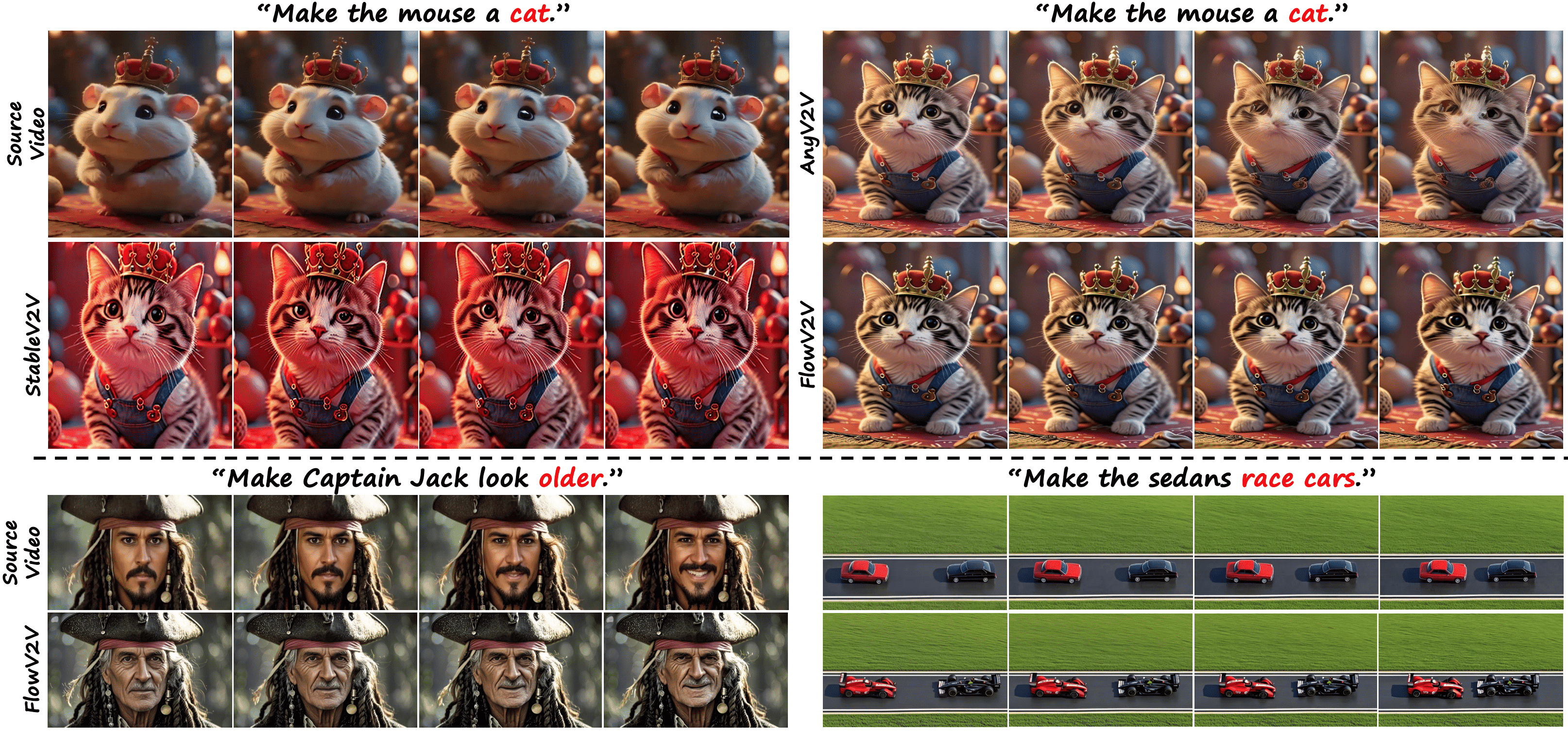}
    \vspace{-1.5em}
    \caption{
Qualitative comparisons of non-rigid motion editing (top), including face rotation, and additional applications (bottom), such as face editing and multi-motion, using our proposed FlowV2V.
    }
    \label{fig:show case}
\end{figure}

\vspace{-0.5em}
\begin{abstract}
With the prosper of video diffusion models, down-stream applications like video editing have been significantly promoted without consuming much computational cost.
One particular challenge in this task lies at the motion transfer process from the source video to the edited one, where it requires the consideration of the shape deformation in between, meanwhile maintaining the temporal consistency in the generated video sequence.
However, existing methods fail to model complicated motion patterns for video editing, and are fundamentally limited to object replacement, where tasks with non-rigid object motions like multi-object and portrait editing are largely neglected.
In this paper, we observe that optical flows offer a promising alternative in complex motion modeling, and present FlowV2V to re-investigate video editing as a task of flow-driven Image-to-Video (I2V) generation.
Specifically, FlowV2V decomposes the entire pipeline into first-frame editing and conditional I2V generation, and simulates pseudo flow sequence that aligns with the deformed shape, thus ensuring the consistency during editing.
Experimental results on DAVIS-EDIT with improvements of $13.67\%$ and $50.66\%$ on DOVER and warping error illustrate the superior temporal consistency and sample quality of FlowV2V compared to existing state-of-the-art ones.
Furthermore, we conduct comprehensive ablation studies to analyze the internal functionalities of the first-frame paradigm and flow alignment in the proposed method.
\end{abstract}

\section{Introduction}
\vspace{-0.7em}
% 第一段
With the exponential growth of digital content, video has become a central medium for daily recording, where this data modality has prompted multiple applications in the field of computer vision, typically including video editing.
To achieve video editing, traditional methods are normally labor-intensive due to the manual operations, and inherently limited in scalability, where automatic solutions are thus expected.
With the development of diffusion models \cite{ho-etal-2020-ddpm,song-etal-2021-ddim,rombach-etal-2022-stable-diffusion,peebles-etal-2023-dit,he2024id,guo-etal-2024-i2vadapter,jain-etal-2023-peekaboo,ardino-etal-2021-click-to-move,zhang-etal-2023-controlnet,ruiz-etal-2023-dreambooth}, video editing has been promoted to a new era, where it is enabled with their strong generative capabilities.

To perform the task of video editing, early-proposed methods \cite{geyer-etal-2024-tokenflow,jay-etal-2023-tuneavideo,liu-etal-2023-videop2p,zi-etal-2024-cococo,cui-etal-2024-stabledrag,feng-etal-2023-ccedits,stablev2v} mainly rely on the motion trajectories of the original video, where they perform the editing process under the assumption that edited objects do not undergo significant shape changes and use techniques like DDIM inversion \cite{geyer-etal-2024-tokenflow,ku-etal-2024-anyv2v,liu-etal-2023-videop2p} and one-shot tuning \cite{liu-etal-2023-videop2p,geyer-etal-2024-tokenflow}.
However, these methods fall short in the scenarios involving significant shape deformation, as is noted by StableV2V \cite{stablev2v}, where in real-world editing cases, users might expect to modify the geometry of objects—particularly in tasks such as multi-object replacement or facial editing—where preserving rigid structure is thus largely neglected in these studies.
Considering the alignment of shape deformation, StableV2V constructs its pipeline based on the first-frame-based methods—which sequentially edits the first video frame and then propagates it to the others through a conditional Image-to-Video (I2V) process—and utilizes depth maps as the intermediate media for motion control.
Yet, depth map inherently struggles to handle complex and non-rigid shape transformations, as it fails to represent motion trajectories when complicated patterns like occlusion, rotation, facial expression changes occur.
More effective condition to represent complex motion patterns is thus expected for the field of video editing.

In this paper, we propose FlowV2V to re-organize video editing as the flow-guided I2V generation process, where we follow the first-frame-based paradigm, and adopt shape-aligned flow for motion control throughout the editing process.
Our method comprises four main components, including First Frame Editing (FFE), Iterative Motion Propagation (IMP), Shape-Consistent Flow Calibration (SCFC), and Flow-Driven Image-to-Video Generation (FD-I2V).
Sequentially, FFE processes prompts of various modalities (e.g., text, instruction, reference image, and so on) and edits the first video frame and latter plays as the initial image condition for the I2V model in FD-I2V.
IMP performs an iterative alignment process, where it obtains a pseudo optical flow sequence with the consideration of shape alignment to the edited objects, and afterward, SCFC further refines the flow sequence to facilitate its preciseness.
Eventually, FD-I2V utilizes a flow-driven I2V model to propagate the contents in first edited frame to all others, and outputs the entire edited video sequence in the manner of I2V generation.
To prove the validity of FlowV2V, we evaluate our method on DAVIS-EDIT \cite{stablev2v}, where results show our superior sample quality, controllability, and flexibility, along with comprehensive ablation studies analyzing the impacts of the modality of control signal and various edited frames.

\vspace{-0.5em}
\section{Related Works}
\vspace{-0.7em}
\paragraph{Video Diffusion Model.}

Diffusion models have achieved remarkable success in a wide series of visual generation tasks \cite{guo-etal-2023-animatediff,ma-etal-2024-latte,cong-etal-2024-flatten,ceylan-etal-2023-pix2video}.
Two particular video-based topics are primarily studied—Text-to-Video (T2V) and Image-to-Video (I2V) generation.
For T2V generation, methods such as WAN \cite{wan} and Hunyuan-DiT \cite{hunyuan} jointly model spatial and temporal dimensions with the network architecture of Diffusion Transformer (DiT) \cite{peebles-etal-2023-dit}.
For I2V generation, I2VGen-XL \cite{zhang-etal-2023-i2vgenxl} and Step-Video\footnote{\url{https://github.com/stepfun-ai/Step-Video-Ti2V}} are two most relevant studies in recent years.
Besides, SVD \cite{blattmann-etal-2023-svd} adopts a spatial-temporal decoupling strategy to balance spatial quality and temporal modeling efficiency, so as to conduct better I2V generation. 
Generally speaking, the aforementioned models enables a wide series of applications, e.g., style transfer, video editing, and etc., especially without tuning task-specific models, where they serve as the fundamental support for future development of the visual generation community.

\vspace{-0.7em}
\paragraph{Video Editing.}

In recent years, video editing has emerged as an important topic.
Existing diffusion-based video editing methods can be broadly categorized into four groups—one-shot tuning methods, DDIM inversion-based methods, learning-based methods, and first-frame-based methods \cite{geyer-etal-2024-tokenflow,jay-etal-2023-tuneavideo,liu-etal-2023-videop2p,zi-etal-2024-cococo,cui-etal-2024-stabledrag,feng-etal-2023-ccedits,song2025univstunifiedframeworktrainingfree,stablev2v}.
Specifically, one-shot tuning methods \cite{liu-etal-2023-videop2p,jay-etal-2023-tuneavideo} adapt model weights for a given video through video-specific tuning, where it enables diverse editing results conditioned on different user-provided text prompts.
DDIM inversion-based methods \cite{geyer-etal-2024-tokenflow,ku-etal-2024-anyv2v,liu-etal-2023-videop2p} reconstruct latent representations by inverting diffusion trajectories, then capture motion patterns and enforcing temporal consistency across generated frames.
Learning-based methods \cite{zhang-etal-2024-avid,zi-etal-2024-cococo} integrate motion modules into pre-trained image diffusion models and tune dedicated video-based inpainting models with a large ammount of video-text pairs.
Compared to all the aforementioned methods, first-frame-based methods \cite{ku-etal-2024-anyv2v,ouyang-etal-2024-i2vedit,stablev2v} offer a flexible paradigm to process various prompts from different modalities, starting by modifying the first frame and propagating the results to subsequent frames via motion trajectory control, with more details provided in the next paragraph.

\vspace{-1.4em}
\paragraph{First-Frame-based Editing Methods.}

This type of methods leverage the modified first frame to drive the entire editing process.
Specifically, one prominent approach is AnyV2V \cite{ku-etal-2024-anyv2v}, which utilizes image editing models to modify the first frame, and then uses an I2V model to propagate the edited contents via temporal feature injection.
Based on AnyV2V, I2VEdit adaptively preserves both the visual and motion integrity of the source video and adjusts the attention maps considering both global and local modifications.
However, due to a lack of effective alignment between the transferred motion and the edited content, video editing cases with significant shape changes normally fail.
To address this issue, StableV2V \cite{stablev2v} aligns the depth map sequence with the edited objects, and establishes alignment between the transferred motion and the user prompts, thus enabling a shape-consistent video editing process.
Even so, depth-map contains its intrinsic problem, where it normally struggle to represent complicated motion patterns like occlusions and rotation, where more effective ways to handle such scenarios are expected, thus promoting our work in this paper.

\vspace{-1.15em}
\section{Methods}
\label{sen_inst}
\vspace{-0.7em}

\begin{figure*}[t]
  \centering
  \includegraphics[width=1\linewidth]{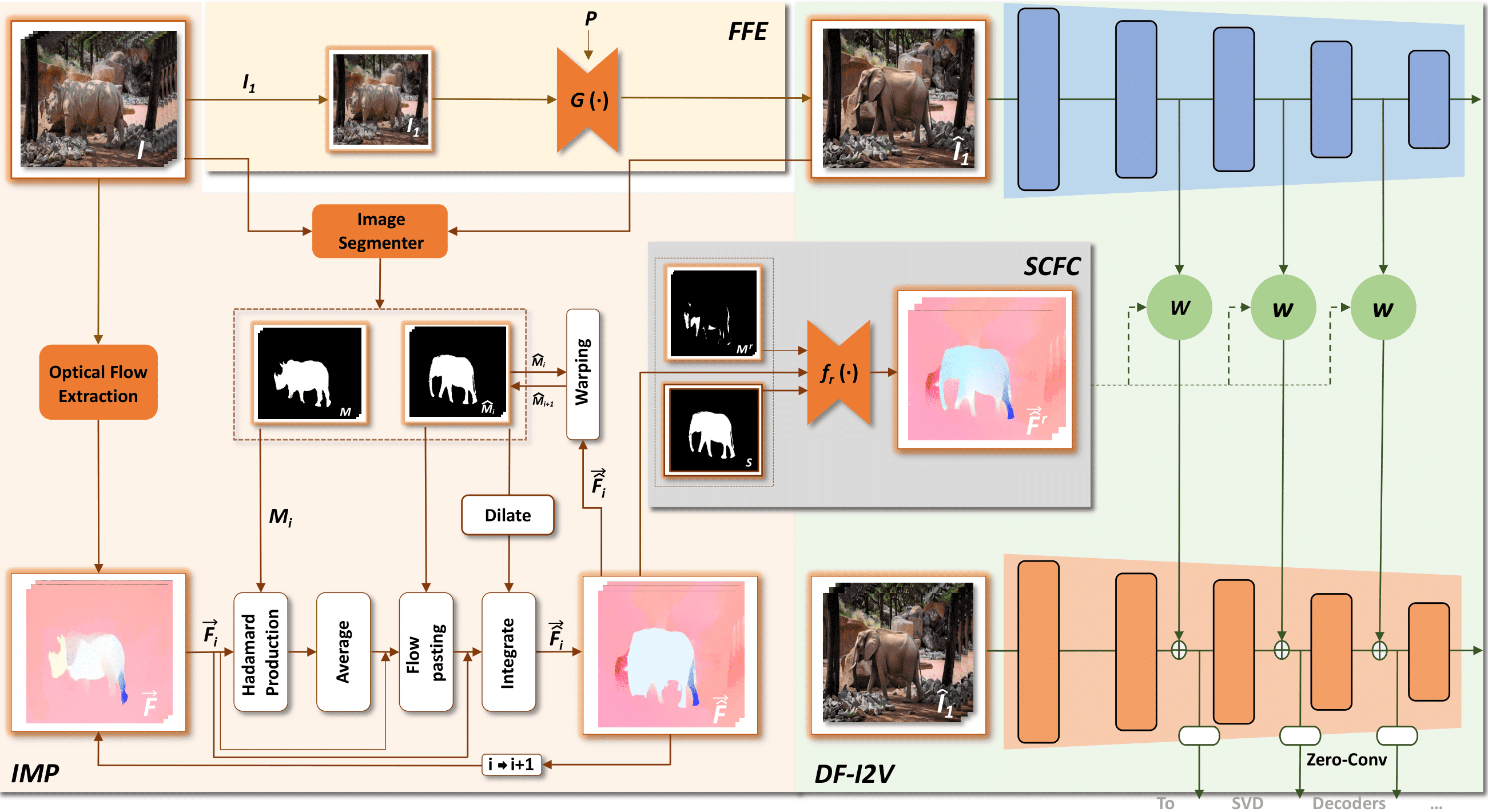}
  \vspace{-2em}
  \caption{
      \textbf{The overall architecture of our proposed FlowV2V video editing pipeline.} It consists of a first-frame editor, an iterative motion propagator, a shape-consistent flow calibration module, a reference encoder capturing multi-scale features from the source image, and a trainable SVD encoder initialized from SVD, serving as the final spatial-temporal feature merger in generation guidance.
  }
  \vspace{-1.7em}
  \label{fig:method}
\end{figure*}

To implement FlowV2V, we follow previous first frame-based methods \cite{ku-etal-2024-anyv2v,ouyang-etal-2024-i2vedit,stablev2v} to decompose the entire video editing pipeline into two parts, i.e., first frame editing and conditional image-to-video generation, where the overview of this work is shown in Figure \ref{fig:method}.
As demonstrated, FlowV2V contains four procedures, including First Frame Editing (FFE), Iterative Motion Propagation (IMP), Shape-Consistent Flow Calibration (SCFC), and Flow-Driven Image-to-Video Generation (FD-I2V).
Specifically, FFE accepts external text or image prompts to add edited contents into the first frame of the input video.
Afterward, IMP extracts the optical flow from the input video, and propagates the averaged motions from the regions of the original object to the ones of edited object, and iteratively perform such operation until all edited frames are processed, resulting in the pseudo optical flow to control I2V generation.
To further ensure the precision of obtained pseudo flow, we draw inspiration from existing video inpainting studies \cite{zhou-etal-2023-propainter} to calibrate redundant motion information in SCFC.
Eventually, we integrate the aforementioned control and transfer the edited contents in the first video frame to all others in a manner of conditional image-to-video generation.
For detailed illustration of the above procedures, we construct our paper in the structures below, with Section \ref{sec:ffe}, \ref{sec:IMP}, \ref{sec:scfc}, and \ref{sec:fdi2v} illustrating FFE, IMP, SCFC, and FD-I2V, respectively.

% Our goal is to generate an edited video that aligns with a user-guided edited first frame, guided by optical flow information extracted from the original video. To this end, we first estimate pseudo optical flow from the original video, and then apply a mask-based iterative strategy along with a flow averaging scheme to obtain a deformed flow that better captures the structural changes of objects in the edited frame. This deformed flow is subsequently fed into an adapter module, which transforms it to be compatible with the SVD network, ultimately enabling the synthesis of the full edited video sequence.

% In the following sections, we describe the key components of our approach in detail. In Section 2.1, we introduce the editing of the first frame image. Section 2.2 covers the extraction of pseudo optical flow from the original video. In Section 2.3, we explain how the deformed flow is obtained using a mask-based iterative strategy and a flow averaging scheme. Section 2.4 describes how to refine the deformed optical flow based on the mask shape. Finally, Section 2.5 details how the adapter receives the optical flow and adjusts it to be compatible with the SVD, thereby completing the video generation process.

\vspace{-0.8em}
\subsection{First Frame Editing}
\label{sec:ffe}
\vspace{-0.5em}

Similar to previous first-frame-based studies \cite{ku-etal-2024-anyv2v,ouyang-etal-2024-i2vedit,stablev2v}, the first step is to process the external prompt into edited contents in the first video frame.
To achieve so, we consider off-the-shelf image editors to process prompts from various modalities accordingly.
Let $P$ denotes the external prompt offered by users, we send the first video frame $I_1 \in V = \{ I_i\}_{i=1}^N$ into the image editor $\mathcal{G}_{img} \left( \cdot \right)$, with $V$ as the input video sequence and $N$ referring to its number of frames, respectively, where we simply write the mathematical formulation of this process by:
\begin{equation}
    \widehat{I_1} = \mathcal{G}_{img}(I_1, P),
\end{equation}
where \(\widehat{I_1}\) represents the edited first frame.
For selection of $G_{img} \left( \cdot \right)$, particularly, we use SD Inpaint \cite{rombach-etal-2022-stable-diffusion}, InstructPix2Pix \cite{brooks-etal-2023-instructpix2pix}, and Paint-by-Example \cite{yang-etal-2023-paint} to handle external prompts in modalities of text, instruction, and reference image, respectively.
For the editing scenarios in the human portrait domain, we adopt a portrait editing toolkit in Jimeng AI\footnote{\url{https://jimeng.jianying.com/ai-tool}}.

\vspace{-0.8em}
\subsection{Iterative Motion Propagation}
\label{sec:IMP}
\vspace{-0.5em}

Once we obtained the edited first frame $\widehat{I}_1$, the next step is to perform the conditional image-to-video generation process and generate the entire edited video.
Previous studies like StableV2V \cite{stablev2v} specifically address the shape misalignment problem in the task of video editing, where it utilizes an iterative alignment algorithm to obtain shape-aligned depth maps as the control signal.
Although such problem is addressed to some extents, we observe that depth maps are often suboptimal to represent motions of moving objects, where depth-based methods are fundamentally limited to the editing cases of rigid body motion.
When handling non-rigid editing scenarios, especially the ones with object occlusions and rotation (as is shown in Figure \ref{fig:show case}), depth-based methods struggle to propagate the edited contents, since depth map normally fails to precisely depict object motions in such cases.
Optical flow seems to be a promising alternative for such problem, yet obtaining feasible flows with shape alignment to the edited object poses a significant challenge, which motivates our proposal of Iterative Motion Propagation (IMP) in this work.
\vspace{-0.7em}
\paragraph{Optical Flow Extraction.}
The first step in IMP is to extract the optical flow from the input video sequence $V$.
Given the $i$-th video frame $I_i \in V = \{ I_i\}_{i=1}^N $, we denote the optical flow considering the direction of $I_i$ to $I_{i+1}$ as $\overrightarrow{F}_i$.
Then, we use a pre-trained optical flow extractor, i.e., RAFT \cite{teed-etal-2020-raft}, which predicts $\overrightarrow{F}_i$ with the inputs of $I_i$ and $I_{i+1}$, with the overall process formulated by:
\vspace{-0.2em}
\begin{equation}
    \overrightarrow{F} =   \{ \overrightarrow{F}_i \}_{i=1}^{N-1} = \{ \mathcal{E}_{flow} \left( I_i, I_{i+1} \right) \}_{i=1}^{N-1}
\end{equation}
\vspace{-0.3em}
where $\mathcal{E}_{flow} \left( \cdot \right)$ denotes the optical flow extractor and $\overrightarrow{F}$ represents the entire optical flow sequence.

\paragraph{Shape-Aware Flow Deformation.}
With the optical flow sequence extracted, we acquire a control signal for the Image-to-Video (I2V) generation model, yet it is not feasible to directly apply $\overrightarrow{F}$ on the I2V model for the reason that there exists shape misalignment between the edited object and the original one.
To alleviate such misalignment, it is vital to deform the original optical flow sequence with consideration of the shape changes, where we propose Shape-Aware Flow Deformation to conduct this process.
Since the edited object and the original one are required to contain consistent motion information under the task of video editing, it is feasible to assume that the average flow values in both regions are theoretically the same.
Therefore in IMP, we utilize the binary mask sequence as an intermediate condition in order to transfer the motion information from the edited regions to the original ones.
In details, we first adopt a pre-trained image segmentation model (i.e., SAM \cite{kirillov-etal-2023-sam}) to obtain a binary mask sequence $M = \{ M_i \}_{i=1}^N$ and the first frame mask $\widehat{M}_1$ from the input video $V$ and the edited first frame $\widehat{I}_1$.
Let $\overrightarrow{F} \left( m, n \right)$ denote the flow value at the $m$-th column and $n$-th row, we compute the average flow value termed $\overline{F}$ within $M$, and assign $\overline{F}$ to each flow point in $\widehat{M}_1$, where the aforementioned processes are written as:
\begin{equation} \label{eq:flow}
    \begin{aligned}
    &M = \{ M_i \}_{i=1}^N = \mathcal{E}_{seg} \left( I_1, \cdots, I_N \right), \\
    &\widehat{M}_1 = \mathcal{E}_{seg} \left( \widehat{I}_1 \right), \\
    &\overline{F}_1 = \frac{1}{\lvert M_1 \rvert} \sum_{\left( m,n \right) \in M_1} \overrightarrow{F} \left( m,n \right), \\
    &\overrightarrow{\widehat{F}}_1 = \overline{F}_1 \  \odot \widehat{M}_1 + \overrightarrow{F}_1 \left( \vmathbb{1} - \widehat{M}_1 \right),
    \end{aligned}
\end{equation}
where $\overrightarrow{\widehat{F}}_1$ represents the pseudo optical flow of the first frame with alignment to the shape of $\widehat{I}_1$; $\odot$ indicates the Hardamard production between matrices ; $\lvert M_1 \rvert$ denotes the number of pixels in $M_1$; $\mathcal{E}_{seg} \left( \cdot \right)$ is the image segmentation model; $\vmathbb{1}$ refers to an all-one matrix.
For further illustration of this process, $\overrightarrow{\widehat{F}_1} \left( m,n \right)$ is the same as the average flow value $\overline{F}_1$ when $\left( m,n \right) \in \widehat{M}$, otherwise equals to $\overrightarrow{F}_1 \left( m,n \right)$.
One challenge herein is that segmentation masks of other frames, i.e., $\{ \widehat{M}_i \}_{i=2}^{N}$, are not available, but we require them to acquire the entire psuedo optical flow sequence $\overrightarrow{\widehat{F}} = \{ \overrightarrow{\widehat{F}}_i \}_{i=1}^{N-1}$.
Based on the binary-valued nature of $\widehat{M}$, we utilize $\overrightarrow{F}_i$ to obtain $\widehat{M}_{i+1}$ through warping $\widehat{M}_i$.
Let $\overrightarrow{F}_i \left( m,n \right) = \left( \mathbf{u}_i \left(m, n \right), \mathbf{v}_i \left( m,n \right) \right)$ represents the displacement of each pixel from its position $\left(m, n \right)$ in $\widehat{M}_i$ to its next-frame position $\left( m + \mathbf{u}_i \left( m,n \right), n + \mathbf{v}_i \left( m,n \right) \right)$ in $\widehat{M}_{i+1}$, where $\mathbf{u}_i$ and $\mathbf{v}_i$ represents two vectors of $i$-th frame representing the motion directions along the height and width dimensions, respectively.
Then, we use $\overrightarrow{F}_i$ to perform the warping operation on $\widehat{M}_i$ as follows:
\begin{equation} \label{eq:warping}
    \widehat{M}_{i+1} \left( m,n \right) = \widehat{M}_i \left( m + \mathbf{u}_i \left( m,n \right), n + \mathbf{v}_i \left( m,n \right) \right).
\end{equation}
By implement Equation \ref{eq:warping} on \ref{eq:flow}, we can obtain the segmentation mask $\widehat{M}_2$ of the second edited frame, which allows us to conduct Equation \ref{eq:flow} based on the $2$-nd frame elements to obtain $\overrightarrow{\widehat{F}}_2$.
Eventually, we repeat the aforementioned iteration with the elements of the $i$-th frame starting from $i=1$ to $i=N-1$, and eventually get the pseudo optical flow sequence $\overrightarrow{\widehat{F}} = \{ \overrightarrow{\widehat{F}} \}_{i=1}^{N-1}$ of the edited video.

\vspace{-0.5em}
\subsection{Shape-Consistent Flow Calibration}
\label{sec:scfc}
\vspace{-0.3em}

With both FFE and IMP finished, we are to offer a control signal for the I2V model to propagate the first-frame edited contents to all others, however in real practices, we observe that such pseudo optical flow may contain redundant regions that tend to misguide the I2V generation process.
This observation particularly occurs when the edited object has non-overlapped regions with the original one, making it an urgent need to improve the preciseness of the control signal by removing these redundant parts in the pseudo optical flow sequence.
On one hand, regarding the requirement of content removal, the task of inpainting fundamentally offers a reference paradigm to solve this requirement, by training neural networks to learn the inpainting process in a self-supervised manner.
From another aspect, we expect to preserve the established shape alignment through the flow calibration process, which thus needs specific condition to guide the refinement networks.
To this end, we draw inspiration from previous flow-based inpainting and editing studies \cite{zhou-etal-2023-propainter,peruzzo-etal-2024-vase} with a tailored network to perform Shape-Consistent Flow Calibration (SCFC) for FlowV2V.
Note that in this paper, we only use the SCFC on the editing cases with obvious shape deformation, otherwise we use the default optical flow from the original video sequence without specified.
\vspace{-0.7em}
\paragraph{Flow Calibration Network.}
We design our flow calibration network for purpose of shape-consistent flow refinement based on the one from ProPainter \cite{zhou-etal-2023-propainter}, where it accepts three inputs, i.e., the corrupted flow sequence $\overrightarrow{F}^m$, the binary mask sequence $M^r$, and the shape guidance $S$.
To obtain the above elements, we first corrupt the ground-truth optical flow $\overrightarrow{F}$ with $M^r$, and then send the concatenation of the resulting flow sequence along with $M^r$ and $S$ into the flow calibration network $\mathcal{G}_{flow} \left( \cdot \right)$, where the aforementioned processes are formulated as the equations below:
\begin{equation} \label{eq:flow-completion}
\begin{aligned}
    &\overrightarrow{F}^m = \overrightarrow{F} \ \odot M^r, \\
    &\overrightarrow{\widehat{F}}^r = \mathcal{G}_{flow} \left(  \overrightarrow{F}^m, M^r, S \right),
\end{aligned}
\end{equation}

where $\overrightarrow{\widehat{F}}^r$ denotes the refined optical flow sequence that is used to control the I2V model afterwards.
Here, we compute two types of $\overrightarrow{F}$ to enhance the temporal modeling ability of the flow calibration network, i.e., forwarding and backward flows, which are termed as $\overrightarrow{F}_f$ and $\overrightarrow{F}_b$, respectively.
In training, we follow the random masking algorithm in FGT \cite{zhang-etal-2022-flowguided} to obtain $M^r$ randomly corrupt $\overrightarrow{F}$.
In inference, we use the pseudo flow sequence $\overrightarrow{\widehat{F}}$ from IMP instead of $\widehat{F}$ in Equation \ref{eq:flow-completion} in order to achieve flow calibration.
As for the shape guidance $S$, we use the segmentation mask $M_1$ of the first video frame $I_1$ in training, and adopt the one (denoted as $\widehat{M}_1$) of edited first frame $\widehat{I}_1$ in inference.
\vspace{-0.7em}
\paragraph{Network Architecture.}
When implementing the flow calibration network, we perform both forward and backward flow estimation processes to achieve best-performing results.
Specifically, we first process the bi-directional flow sequences with down-sampling blocks with a ratio of $8$, and then employ several deformable convolution layers to propagate flow information bi-directionally between adjacent frames.
Here, we warp the down-sampled features with its optical flow and leverage a stack of convolutional layers to compute the offsets from deformable convolutions.
Finally, we adopt a CNN-based decoder to predict the final optical flow sequence.
\vspace{-0.7em}
\paragraph{Optimization Objective.}
To optimize the flow calibration network, we compute the reconstruction loss $\mathcal{L}_{rec}$ through the L1 norm-based Euclidean distance between estimated flow sequence and the ground-truth one.
Besides, we adopt a second-order smoothness loss $\mathcal{L}_{smooth}$ following UnFlow \cite{unflow} to promote the smoothness of neighbor regions in the estimated flow.

\vspace{-1em}
\subsection{Flow-Driven Image-to-Video Generation}
\vspace{-0.5em}
\label{sec:fdi2v}

Once the calibrated flow sequence is obtained, we are finally able to produce the final edited video.
In details, we use Stable Video Diffusion (SVD) \cite{blattmann-etal-2023-svd} as our foundation generative model, and insert flow-based adapter layers into SVD, similar to the ones in MOFA-Video \cite{mofa-video}.
To produce the final edited video $\widehat{V} = \{ \widehat{I} \}_{i=1}^N$, we send the edited first frame $\widehat{I}_1$ and the refined optical flow sequence $\overrightarrow{\widehat{F}}^r$ into SVD to propagate the edited contents in $\widehat{I}_1$ to all others in a conditional I2V manner.
\vspace{-0.7em}
\paragraph{Hierarchical Feature Warping and Aggregation.}
The flow-based adapter composes a reference encoder and a feature fusion encoder, where the former adopts multi-scale convolutional blocks to extract hierarchical features from the input first frame; the latter warps the extracted features and fuses them to the corresponding layers in a duplicated SVD encoder network, following a similar strategy of ControlNet \cite{zhang-etal-2023-controlnet}.
Then, the outputted features of the duplicated SVD encoder are aggregated to the SVD decoder, and control the I2V process in producing a motion-aligned edited video sequence.

\begin{table*}[t]
    \centering
     \caption{Quantitative comparison on DAVIS-EDIT \cite{stablev2v} with respect to DOVER \cite{wu-etal-2023-dover}, FVD \cite{thomas-etal-2019-fvd}, WE, CLIP-Temporal, and CLIP score \cite{radford-etal-2021-clip}. We highlight the scores tested under the original height-width ratio with ``$*$''. $\bar{T}$ represents the average inference time per example in minutes across both DAVIS-EDIT-S and DAVIS-EDIT-C. The optimal and suboptimal results are in \textcolor{red}{\textbf{red}} and \textcolor{blue}{\textbf{blue}}, respectively.}
    \scalebox{0.72}{
        \begin{tabular}{l ccccc ccccc c}
            \toprule
            \multirow{2}{*}{\textbf{Method}} & \multicolumn{5}{c}{\textbf{DAVIS-Edit-S}} & \multicolumn{5}{c}{\textbf{DAVIS-Edit-C}} & \multirow{2}{*}{\textbf{$\bar{T}$$\downarrow$}} \\
            \cmidrule(lr){2-6} \cmidrule(lr){7-11}
            & DOVER $\uparrow$ & FVD $\downarrow$ & WE $\downarrow$ & CLIP-T $\uparrow$ & CLIP-S $\uparrow$ & DOVER $\uparrow$ & FVD $\downarrow$ & WE $\downarrow$ & CLIP-T $\uparrow$ & CLIP-S $\uparrow$ & \\
            \midrule
            \rowcolor{RowColor} \multicolumn{12}{c}{\textit{Text-based Evaluation Settings}} \\
            \midrule
            TokenFlow & 66.36 & 17.33 & 18.58 & 95.84 & 24.89 & 67.47 & 17.45 & 18.60 & 95.61 & 24.12 & 5.81 \\
            DMT & 59.27 & 19.53 & 16.65 & 94.11 & 24.91 & 57.45 & 21.64 & 19.89 & 93.58 & 24.51 & 8.88 \\
            I2VEdit & 69.63 & 15.63 & 16.94 & 96.18 & 24.99 & 69.57 & 17.62 & 16.16 & 96.50 & 24.92 & 10.25 \\
            AnyV2V & 66.82 & \textcolor{blue}{\textbf{14.87}} & 15.35 & 95.66 & 25.09 & 65.01 & 17.83 & 18.26 & 94.36 & 24.32 & 8.28 \\
            StableV2V & 67.78 & \textcolor{red}{\textbf{13.77}} & 15.95 & 96.34 & \textcolor{red}{\textbf{25.46}} & 70.80 & \textcolor{red}{\textbf{17.18}} & 15.27 & 96.83 & \textcolor{red}{\textbf{25.68}} & \textcolor{blue}{\textbf{3.14}} \\
            \midrule
            \rowcolor{gray!10} \textbf{FlowV2V} & \textcolor{red}{\textbf{77.20}} & 15.56 & \textcolor{red}{\textbf{5.61}} & \textcolor{red}{\textbf{97.05}} & \textcolor{blue}{\textbf{25.17}} & \textcolor{red}{\textbf{76.96}} & \textcolor{blue}{\textbf{17.28}} & \textcolor{blue}{\textbf{11.43}} & \textcolor{red}{\textbf{96.94}} & 24.98 & \textcolor{red}{\textbf{2.01}} \\
            \rowcolor{gray!20} \textbf{FlowV2V$^*$} & \textcolor{blue}{\textbf{71.77}} & 17.10 & \textcolor{blue}{\textbf{9.31}} & \textcolor{blue}{\textbf{96.99}} & 24.20 & \textcolor{blue}{\textbf{72.49}} & 18.37 & \textcolor{red}{\textbf{9.54}} & \textcolor{blue}{\textbf{96.93}} & \textcolor{blue}{\textbf{25.43}} & 4.26 \\
            \midrule
            \rowcolor{RowColor} \multicolumn{12}{c}{\textit{Image-based Evaluation Settings}} \\
            \midrule
            AnyV2V & 65.83 & \textcolor{blue}{\textbf{12.97}} & 24.47 & 95.89 & \textcolor{blue}{\textbf{25.41}} & 64.56 & 15.25 & 25.61 & 96.13 & \textcolor{red}{\textbf{24.79}} & 8.43 \\
            StableV2V & 67.58 & \textcolor{red}{\textbf{12.36}} & 22.17 & 96.51 & 26.24 & 68.42 & \textcolor{blue}{\textbf{14.87}} & 21.23 & 96.71 & 26.55 & \textcolor{blue}{\textbf{3.23}} \\
            \midrule
            \rowcolor{gray!10} \textbf{FlowV2V} & \textcolor{red}{\textbf{80.62}} & 13.98 & \textcolor{blue}{\textbf{14.88}} & \textcolor{blue}{\textbf{97.18}} & 25.51 & \textcolor{red}{\textbf{77.18}} & \textcolor{red}{\textbf{14.63}} & \textcolor{red}{\textbf{6.02}} & \textcolor{blue}{\textbf{96.74}} & \textcolor{blue}{\textbf{24.83}} & \textcolor{red}{\textbf{2.23}} \\
            \rowcolor{gray!20} \textbf{FlowV2V$^*$} & \textcolor{blue}{\textbf{76.36}} & 17.06 & \textcolor{red}{\textbf{11.41}} & \textcolor{red}{\textbf{97.19}} & \textcolor{red}{\textbf{24.97}} & \textcolor{blue}{\textbf{70.32}} & 16.63 & \textcolor{blue}{\textbf{14.74}} & \textcolor{red}{\textbf{96.84}} & 24.98 & 4.57 \\
            \bottomrule
        \end{tabular}
    }
    \vspace{-1em}
    \label{tab:example}
\end{table*}

\vspace{-0.7em}
\section{Implementation Details}
\vspace{-1em}
\paragraph{Evaluation Setting.}
We follow the evaluation setting of in StableV2V \cite{stablev2v}, where we utilize its proposed DAVIS-EDIT dataset.
Specifically, the dataset consists of two subsets to address editing cases with Similar (S) and Changing (C) shape deformation, i.e., DAVIS-EDIT-S and DAVIS-EDIT-C, respectively.
In total, it contains $50$ editing instances, with $28$ and $22$ videos for text-based and image-based evaluations, where we employ its official annotations.
\vspace{-1em}
\paragraph{Experiment Setting.}
We employ RAFT \cite{teed-etal-2020-raft} for optical flow estimation, using 20 inference iterations to balance accuracy and computation. Following the estimated flows, we synthesize video frames using pre-trained text-to-image diffusion models. However, due to the resolution constraints of these models (mostly based on Stable Diffusion \cite{rombach-etal-2022-stable-diffusion}), we primarily generate video frames under the resolution of $512\times 512$, where notably, StableV2V also follows this setting.
Yet, we discover that videos from DAVIS \cite{ponttuset-etal-2018-davis} (the data source of DAVIS-Edit) contain a certain height-width ratio, where the setting of $512\times 512$ resolution fundamentally compress the video to an unnatural size.
Therefore, we adopt both the default setting (with height-width ratio of $1:1$) as well as the original ratio when conducting video editing, where quantitative results under the latter setting are highlighted with $*$.

\begin{figure*}[t]
    \centering
    \includegraphics[width=1\linewidth]{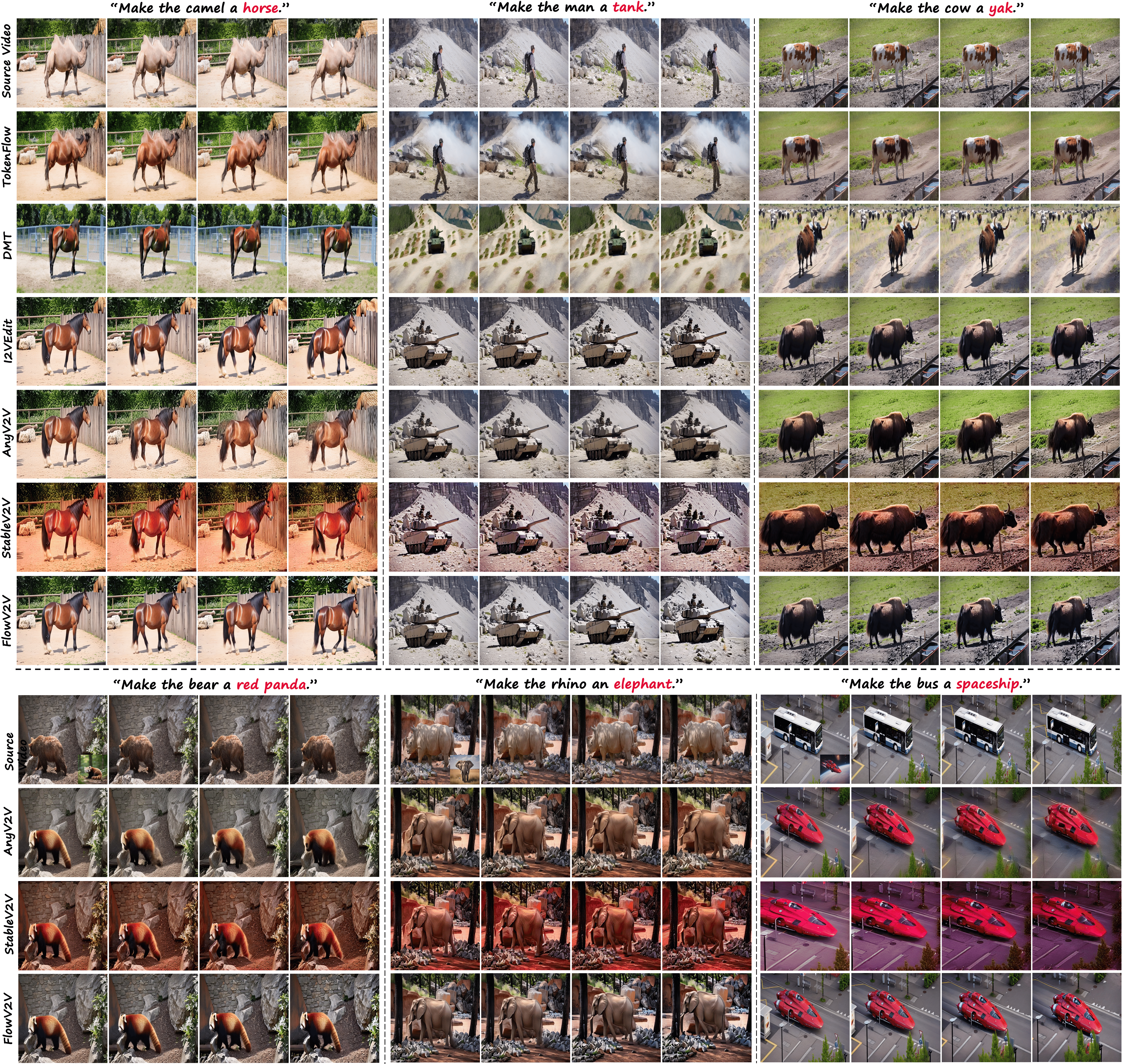}
    \vspace{-1.5em}
    \caption{
        Qualitative comparison of FlowV2V with state-of-the-art methods.
    }
    \vspace{-1.5em}
    \label{fig: comparison}
  \end{figure*}

\vspace{-1em}
\paragraph{Evaluation Metrics.}
We benchmark all methods along three dimensions, i.e., visual fidelity, temporal coherence, text-image alignment, where we adopt DOVER, FVD, inter-frame Warping Error (WE), CLIP-Temporal (CLIP-T), and image-text CLIP score.
Besides, we conduct a user study to measure the human preferences of generated results by different methods.
\vspace{-1em}
\section{Experiment}
\label{headings}

\vspace{-0.7em}
\subsection{Comparison with Other State-of-the-Art Methods}
\vspace{-0.5em}
In this section, we demonstrate and compare FlowV2V against several state-of-the-art video editing approaches, including StableV2V, AnyV2V, I2VEdit, TokenFlow and DMT. 

\vspace{-0.7em}
\paragraph{Performance Comparison.}
Table \ref{tab:example} presents the quantitative evaluation of FlowV2V against existing methods on the DAVIS-EDIT \cite{stablev2v}.
Here,  TokenFlow \cite{geyer-etal-2024-tokenflow}
and DMT \cite{danah-etal-2024-dmt} exhibit deficiencies in maintaining fidelity of the original video and obtain inferior temporal consistency on WE and CLIP-Temporal.
Relatively, I2Edit, AnyV2V and StableV2V achieve improved performance, yet the visual quality of their results are still limited.
In contrast, FlowV2V demonstrates superior performance upon all evaluated metrics, particularly achieving outstanding sample quality and temporal consistency on DOVER, WE, and CLIP-Temporal, meanwhile preserving comparable performance on the other metrics, where this comparison confirms the effectiveness of our flow-driven editing paradigm.
Notably, StableV2V points out that most studies present poorer on DAVIS-Edit-C rather than DAVIS-Edit-S, since the editing cases with significant shape changes are intrinsically more challenging, where FlowV2V performs consistently owing to the stable control offered by the aligned pseudo flow.

% \paragraph{Human Evaluation.}

% We further perform user studies to assess the effectiveness of the proposed approach against baseline methods. We conducted a user study with 13 participants to evaluate the perceptual quality of edited videos. Each participant was shown the input video, user prompt, and results generated by different methods on 22 cases (including 10 from DAVIS-EDIT-S and 12 from DAVIS-EDIT-C, evenly split between image-based and text-based prompts). They were asked to rank the video results by perceptual quality without being informed of the corresponding methods. The average ranks for each method are reported in \textcolor{red}{Table ??}. A Friedman test revealed a significant difference among the methods (p< 0.01). Nemenyi post-hoc analysis confirmed that OURS significantly outperforms other methods (p < 0.05).

\begin{figure*}[t]
  \centering
  \includegraphics[width=1.16\linewidth]{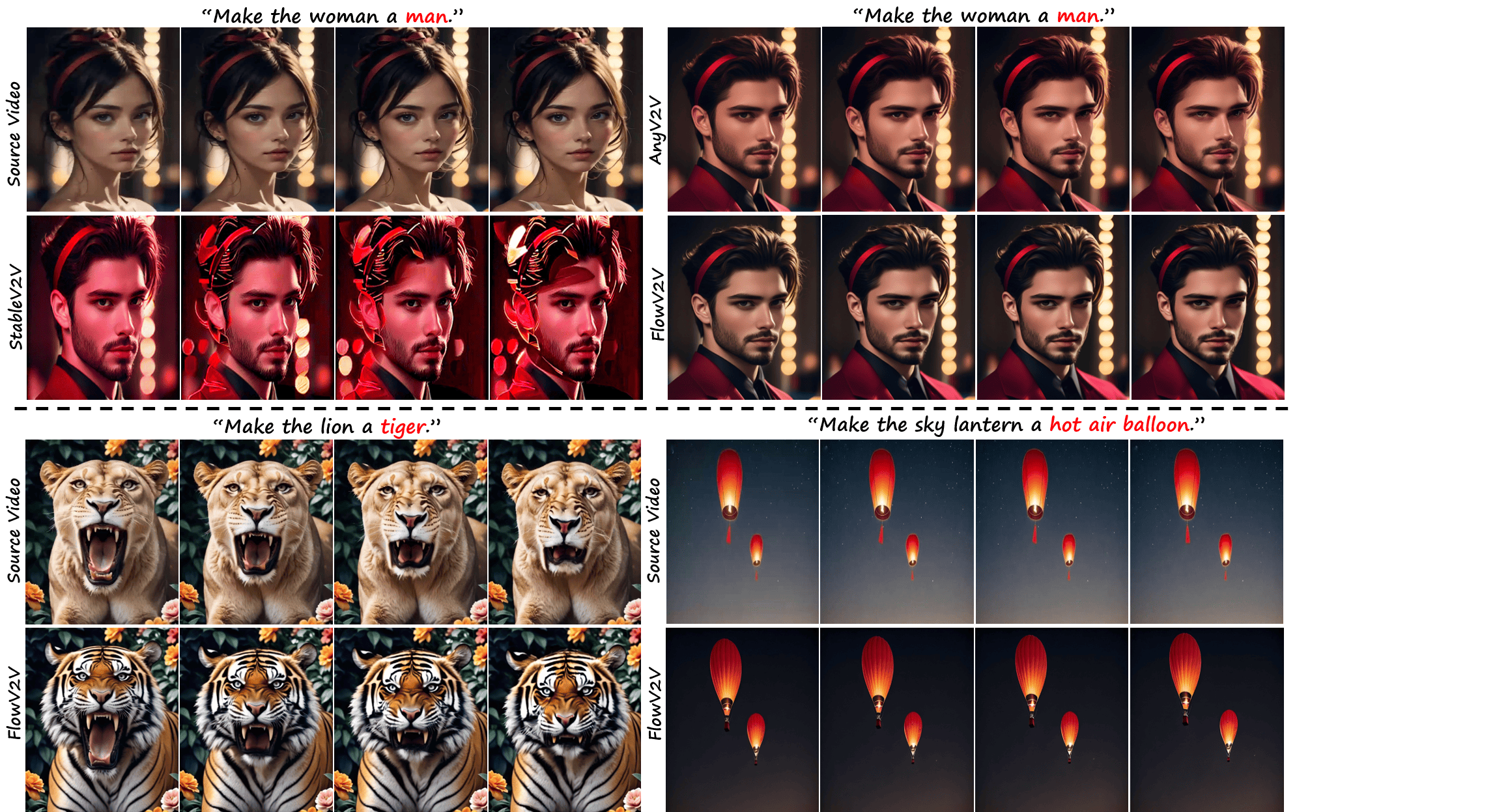}
  \vspace{-1.8em}
  \caption{
      Human portrait and multi-object editing results with FlowV2V.
  }
  \vspace{-0.8em}
  \label{fig: applications}
\end{figure*}

\vspace{-1em}
\paragraph{Applications.}

Despite of the classical usage shown in Figure \ref{fig: comparison}, FlowV2V demonstrates superior on other challenging applications with complicated motion patterns, where the results are illustrated in Figure \ref{fig: applications}.
Here, we particularly compare FlowV2V with depth-guided (StableV2V) and latent features-guided methods (AnyV2V), to show the advantages of optical flow under challenging scenarios.
In Figure \ref{fig: applications}, StableV2V and AnyV2V produce still contents when editing human portrait, since both depth map sequence and latent features are fundamentally limited to model such complex motion patterns.
Conversely, FlowV2V consistently edits all frames, especially when shape deformation occurs, confirming that the superior ability and robustness of our method.
Similar phenomenon is observed under multi-object editing, where the compared two methods produce videos with significant artifacts, and in contrast, such observation is eliminated in our results.

% \begin{table}
% \centering
% \begin{tabular}{llll} 
% \hline
% Method    & D.-E.-S & D.-E.-C & Avg.  \\ 
% \hline
% TokenFlow & 1       & 1       & 1     \\
% DMT       & 1       & 1       & 1     \\
% I2VEdit   & 1       & 1       & 1     \\
% AnyV2V    & 1       & 1       & 1     \\
% Stabl2V2V & 1       & 1       & 1     \\
% OURS      & 1       & 1       & 1     \\
% \hline
% \end{tabular}
% \parbox{\textwidth}{\footnotesize \textcolor{red}{Table??}: This is a long note that will automatically wrap to the next line if it exceeds the table width. You can write multiple sentences here.}
% \end{table}

\begin{figure*}[t]
  \centering
  \includegraphics[width=1\linewidth]{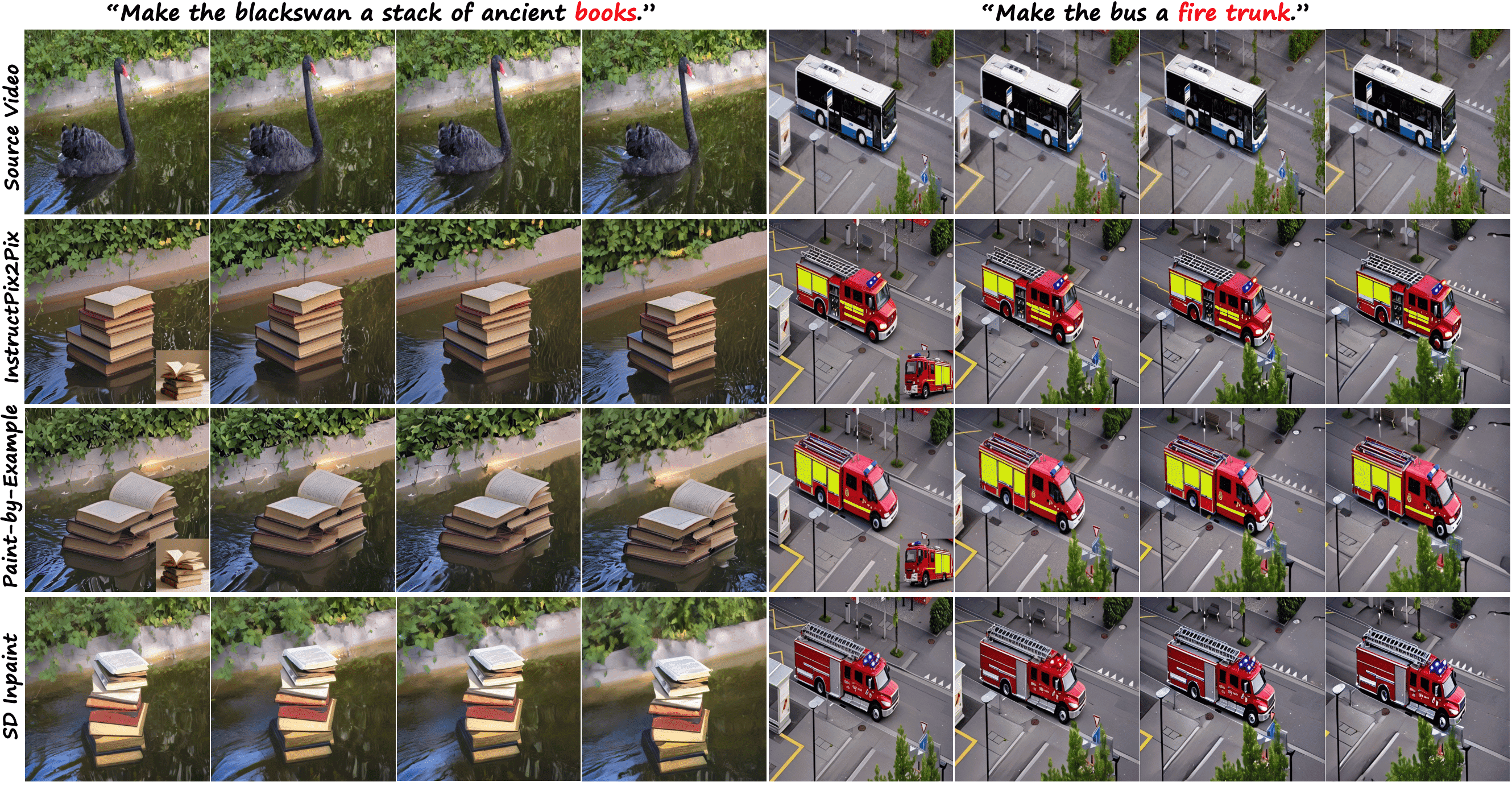}
  \vspace{-1.8em}
  \caption{
      Qualitative results for the ablation study of different first-frame inputs.
  }
  \vspace{-1em}
  \label{fig: first-frame-ablation}
\end{figure*}

\begin{figure*}[t]
  \centering
  \includegraphics[width=1\linewidth]{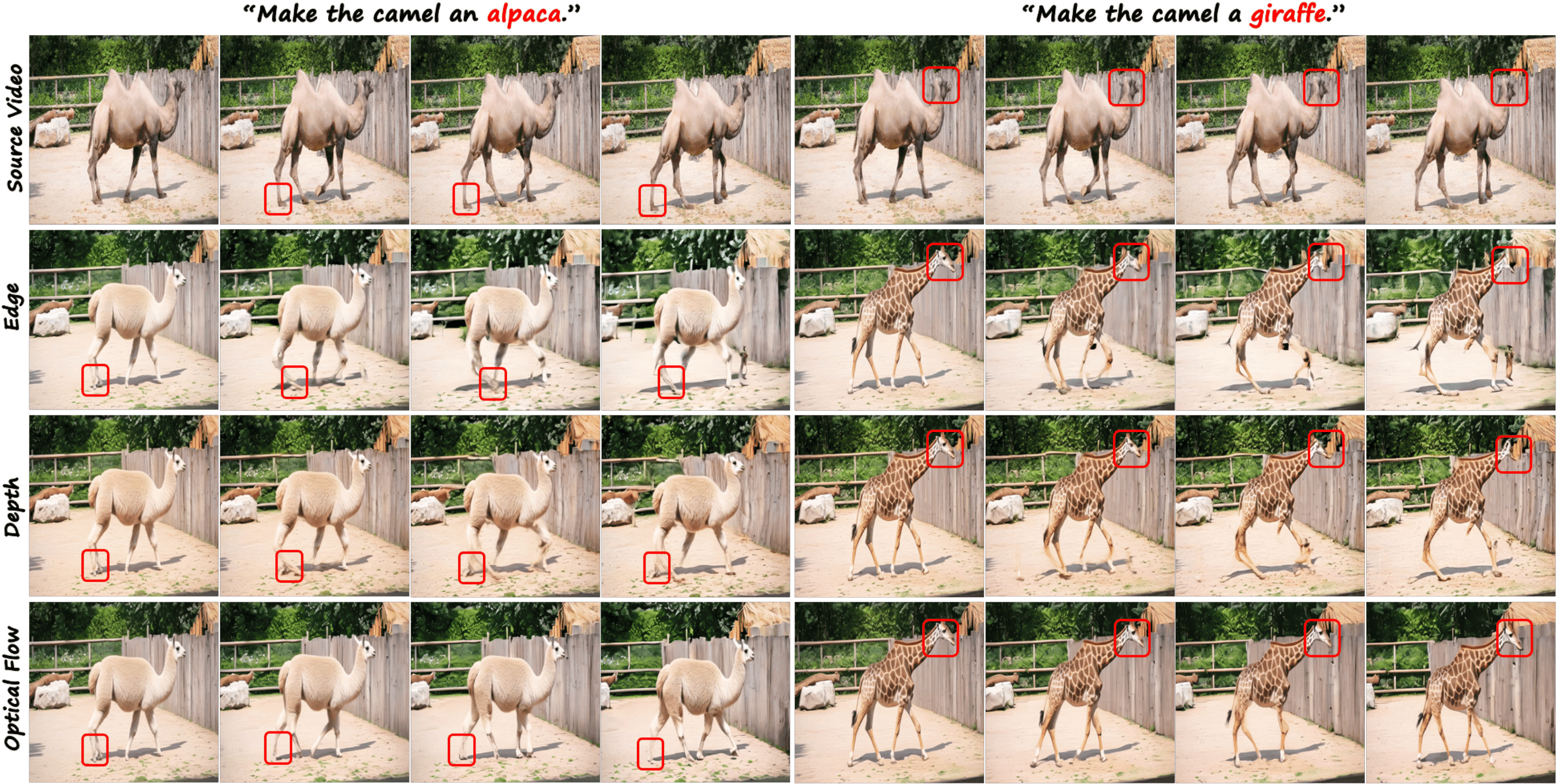}
  \vspace{-1.8em}
  \caption{
      Qualitative results for the ablation study of different condition types.
  }
  \vspace{-1.5em}
  \label{fig: condition-type}
\end{figure*}

\vspace{-0.5em}
\subsection{Ablation Studies}
\vspace{-0.5em}
\paragraph{Evaluating First Frame Editors for Video Editing.}

We evaluated the impact of various first-frame editors on FlowV2V, where results are presented in Figure \ref{fig: first-frame-ablation}.
Specifically, we applied SD Inpaint \cite{rombach-etal-2022-stable-diffusion}, Paint-by-Example \cite{yang-etal-2023-paint}, InstructPix2Pix \cite{brooks-etal-2023-instructpix2pix} to explore the impacts of text, image, and instruction prompts, respectively.
Although different editors exhibit some visual differences in the edited frames, the overall quality of the final edited videos remains consistent, with no significant degradation or variation caused by the selection of editor.
It is worth mentioning that aligning the motions offers a potential improvement on visual quality, when editing a bus into a fire ambulance with SD Inpaint, where FlowV2V produces plausible with photorealistic details like flashing lights.
\vspace{-1em}
\paragraph{Evaluating the Impact of Different Control Signals.}

To verify the effectiveness of our optical flow-based conditioning strategy in the I2V task, we compare two different variants for the optical flow: depth and edge, as shown in Figure \ref{fig: condition-type}.
Here, we highlight the local details by different methods so as to 
straightforward show their effects.
Specifically, in scenarios involving relatively small deformations, using depth or edge as control inputs often leads to object dragging artifacts and motion blur in certain regions of the generated videos.
As the deformation becomes more severe, these artifacts become increasingly prominent, affecting larger areas and further degrading the video quality. 
For further illustration, depth and edge serve as two extreme cases in offering control signal for I2V models, where the former condition is natively coarse for fine-grained motion modeling; the latter condition might cause over-control problem due to its magnitude of control is too much.

\vspace{-1em}
\section{Conclusion}
\vspace{-0.5em}

In this paper, we propose FlowV2V, a novel method to perform video editing through a flow-driven image-to-video generation process, along with a shape-aware flow alignment and a refinement procedures to ensure the preciseness of propagated motions during editing.
Experiments on DAVIS-Edit validates the effectiveness and superior sample quality of FlowV2V, with further ablation studies offering more potential insights of it.
Yet, we observe that FlowV2V might struggle to handle some editing scenarios, since its performance upper bound is fundamentally restricted by the generative power of the used I2V backbone, where we expect to investigate more possible applications with FlowV2V as stronger foundation models emerge in the future.

% \section*{References}
\newpage
\medskip

\bibliographystyle{plainnat}
\bibliography{Styles/neurips_2025}

%%%%%%%%%%%%%%%%%%%%%%%%%%%%%%%%%%%%%%%%%%%%%%%%%%%%%%%%%%%%
\appendix
\newpage
\section*{Technical Appendices and Supplementary Material}
\vspace{-0.9em}
In our supplementary materials, we present additional details and results of FlowV2V to provide deeper insights into the proposed method. The contents are organized according to the following structure:

\vspace{-0.6em}

\begin{itemize}
    \item[$\bullet$] \textbf{Section \hyperref[sec:imp]{A}}: Implementation Details of the Flow Calibration Network.
    
    \item[$\bullet$] \textbf{Section \hyperref[sec:sen_inst]{B}}: Implementation Details of DAVIS-EDIT.

    \item[$\bullet$] \textbf{Section \hyperref[sec:more_results]{C}}: More Results.

    \item[$\bullet$] \textbf{Section \hyperref[sec:raft]{D}}: Impact of RAFT's Estimation Error.

    \item[$\bullet$] \textbf{Section \hyperref[sec:efficiency]{E}}: Efficiency Study.

    \item[$\bullet$] \textbf{Section \hyperref[sec:limitations]{F}}: Limitations.
    
\end{itemize}

\vspace{-0.6em}

\vspace{-0.2em}
\subsection*{A. Implementation Details of Flow Calibration Network}
\label{sec:imp}
\vspace{-0.3em}
This section details the implementation of the flow calibration network, covering its motivation, architecture, and training strategy.
\vspace{-0.8em}
\paragraph{Network Architecture.}
The flow calibration network is a critical component in the FlowV2V system, as its performance directly affects the accuracy of flow guidance in the DF-I2V module, thereby influencing the overall consistency of the edited video.
The primary objective of this network is to remove redundant regions in the input optical flow maps and ensure that the refined flow maps remain consistent with the edited first frame, ultimately enhancing the quality of video editing results. To achieve this, we draw inspiration from ProPainter \cite{zhou-etal-2023-propainter}.
\vspace{-0.8em}
\paragraph{Training Details.}
We train the flow calibration network on the YouTube-VOS dataset, whose training set contains 3,471 videos along with their corresponding segmentation mask annotations. 
We use RAFT \cite{teed-etal-2020-raft} to extract optical flow maps for our approach. 
After data preprocessing, the network is trained for 500k iterations with a batch size of 8. 
Specifically, in each training iteration, we randomly sample 10 optical flow frames and apply the random mask generation algorithm from flow-guided transformer. 
The AdamW optimizer is employed to update the model parameters, with the initial learning rate set to $10^{-4}$.
We implement our method using the PyTorch framework and train it on 4 H20-NVLink (96GB) GPUs.

\vspace{-0.8em}
\subsection*{B. Implementation Details of the DAVIS-EDIT}
\label{sec:sen_inst}
\vspace{-0.3em}
In this section, we provide further details on the implementation of the DAVIS-EDIT benchmark. 
DAVIS-EDIT serves as a crucial component for evaluating the performance of FlowV2V and is carefully constructed by the StableV2V \cite{stablev2v} team to establish a standardized protocol for addressing the shape misalignment problem in video editing. 
To generate the text prompts, the team selectively modifies specific words describing the primary elements in the video, such as objects and foregrounds, with a particular focus on highlighting shape discrepancies during annotation. For instance, “blackswan” is replaced with “duck” to represent cases with similar object shapes, and subsequently, “duck” is replaced with “rabbit” to simulate scenarios involving shape variation. 
The annotation of reference images follows a similar principle, emphasizing the diversity of shape differences. Additionally, the StableV2V team curated reference images that are difficult to be precisely described by text alone, such as the Transformer truck, to further underscore the importance of image-based guidance in these challenging cases.

\vspace{-0.2em}
\subsection*{C. More Results}
\label{sec:more_results}
\begin{figure}[]
  \centering
  \includegraphics[width=\textwidth]{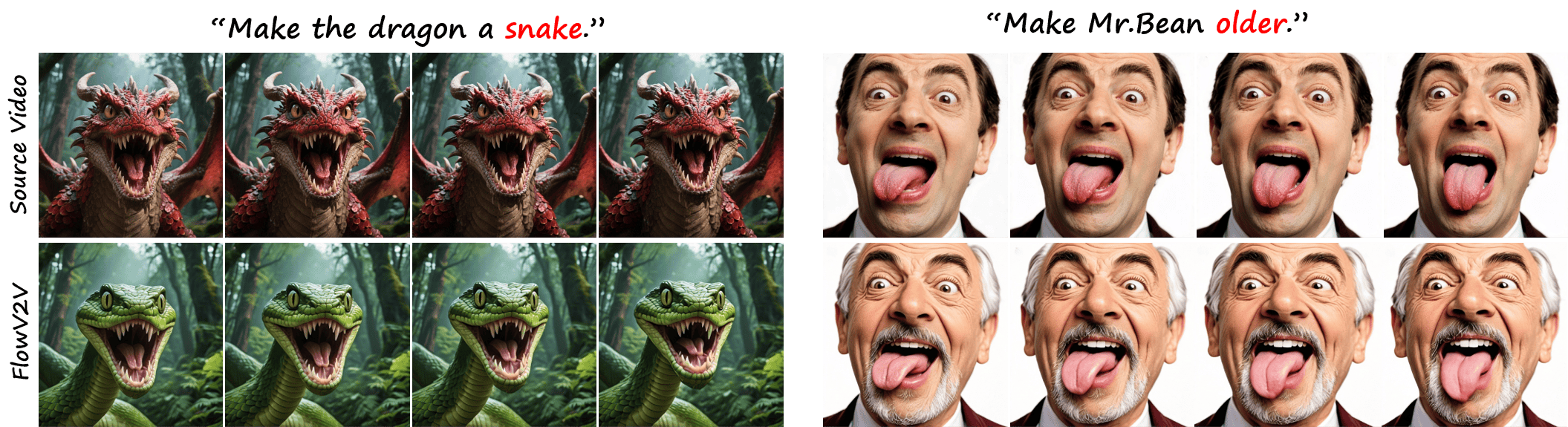}
  \vspace{-1.5em}
  \caption{
      More results under the facial image editing scenario.
  }
  % \vspace{-0.8em}
  \label{fig: face-editing}
\end{figure}

\begin{figure}[]
  \centering
  \includegraphics[width=\textwidth]{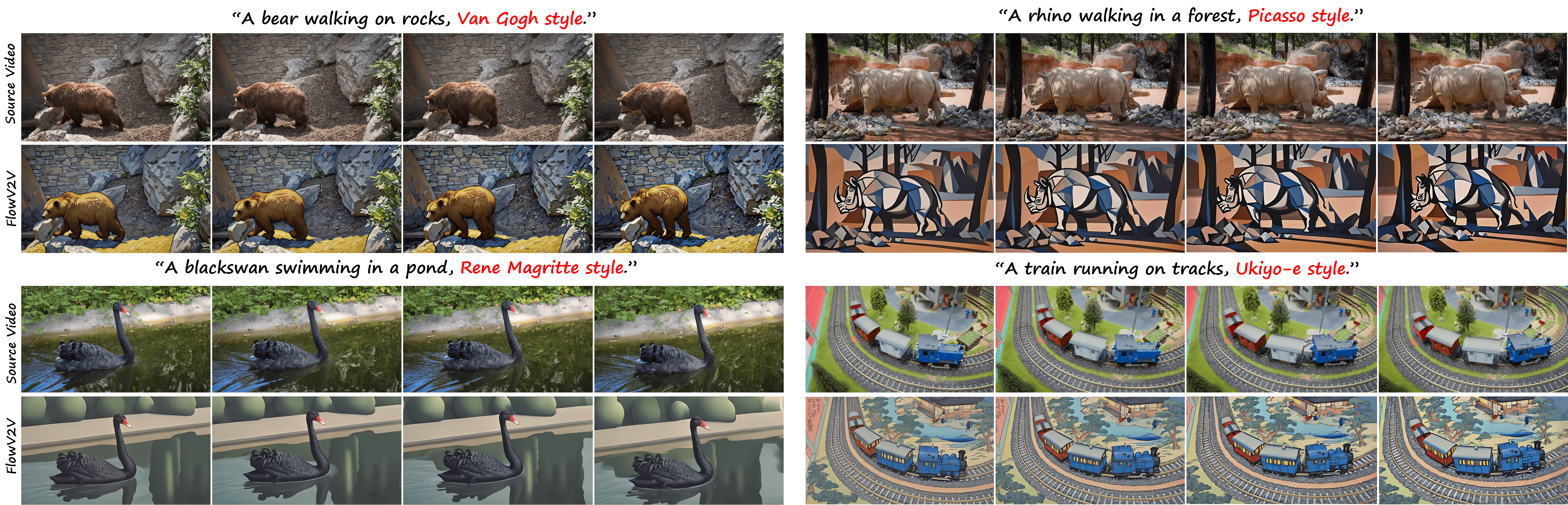}
  \vspace{-1.5em}
  \caption{
      More results under the video style transfer scenario.
  }
  % \vspace{-0.8em}
  \label{fig: style-transfer}
\end{figure}

\begin{figure}[]
  \centering
  \includegraphics[width=\textwidth]{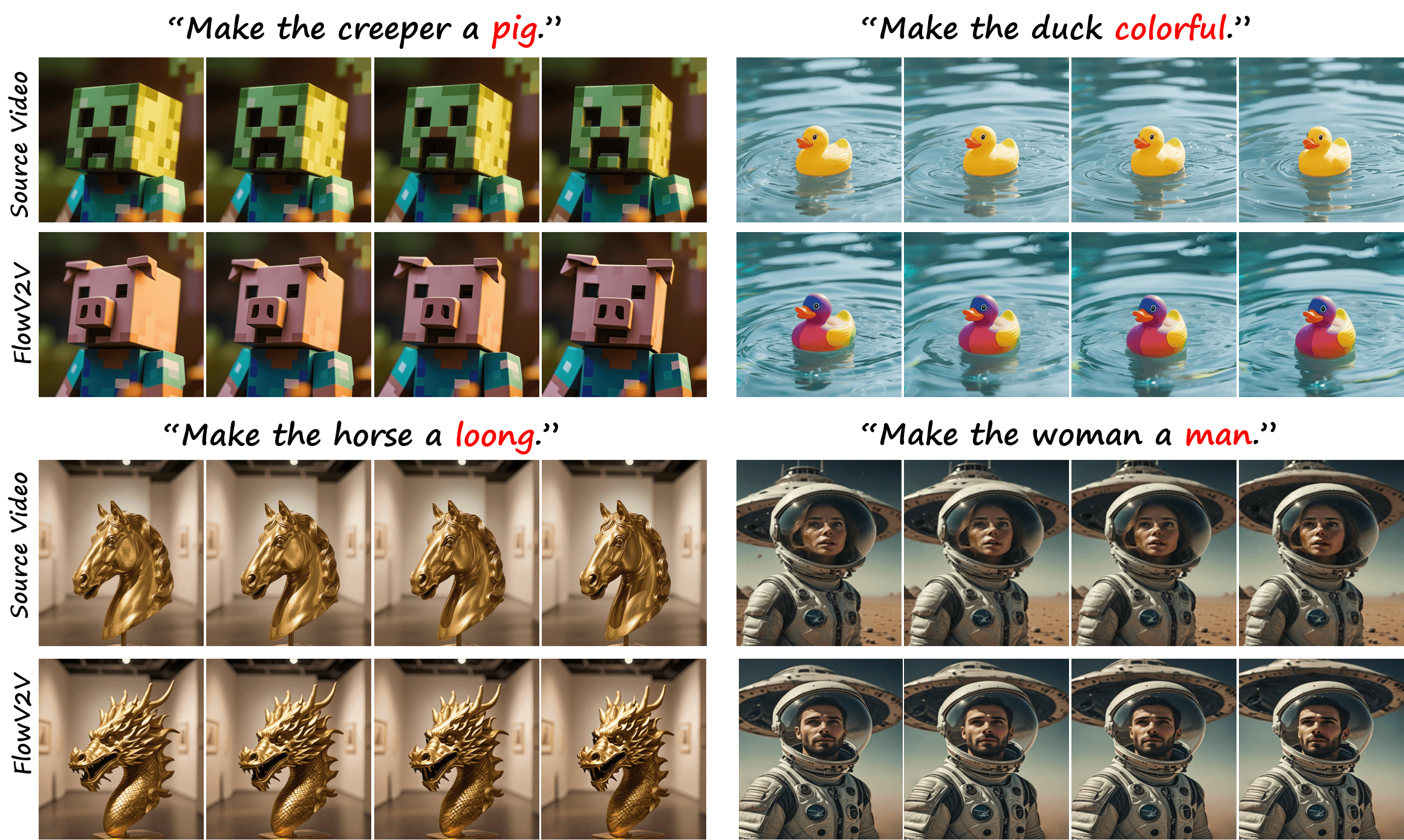}
  \vspace{-1.5em}
  \caption{
      More results under the rotation-based editing scenario.
  }
  % \vspace{-0.8em}
  \label{fig: rotation}
\end{figure}

\begin{figure}[]
  \centering
  \includegraphics[width=\textwidth]{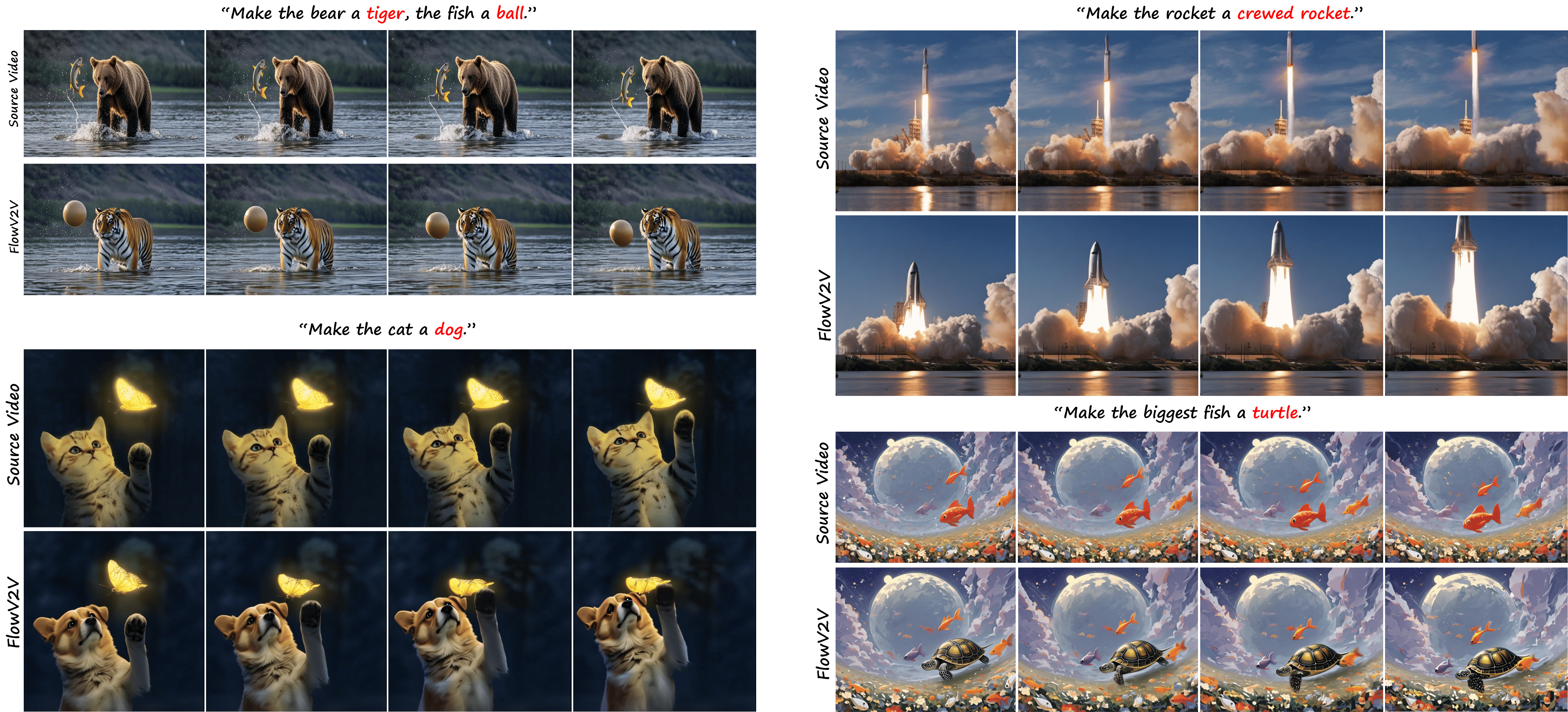}
  \vspace{-1.5em}
  \caption{
      More results under the multi-object editing scenario.
  }
  % \vspace{-0.8em}
  \label{fig: multi-object}
\end{figure}

\begin{figure}[]
  \centering
  \includegraphics[width=\textwidth]{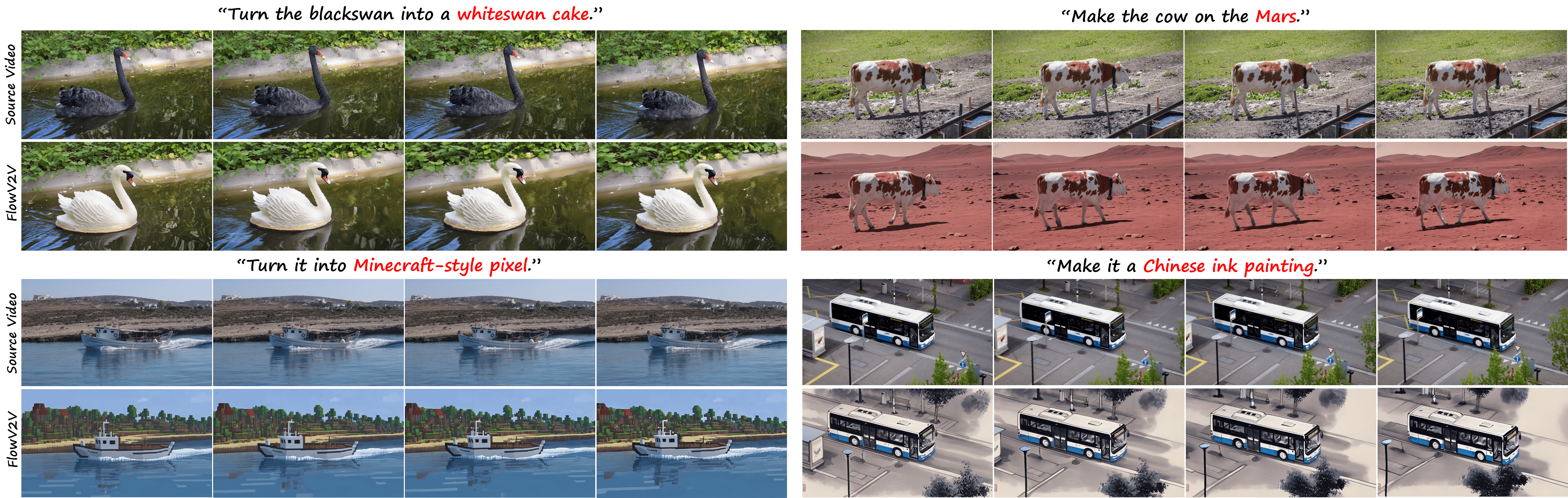}
  \vspace{-1.5em}
  \caption{
      More results under the instruction-based editing scenario.
  }
  % \vspace{-0.8em}
  \label{fig: instruction-based}
\end{figure}

\begin{figure}[]
  \centering
  \includegraphics[width=\textwidth]{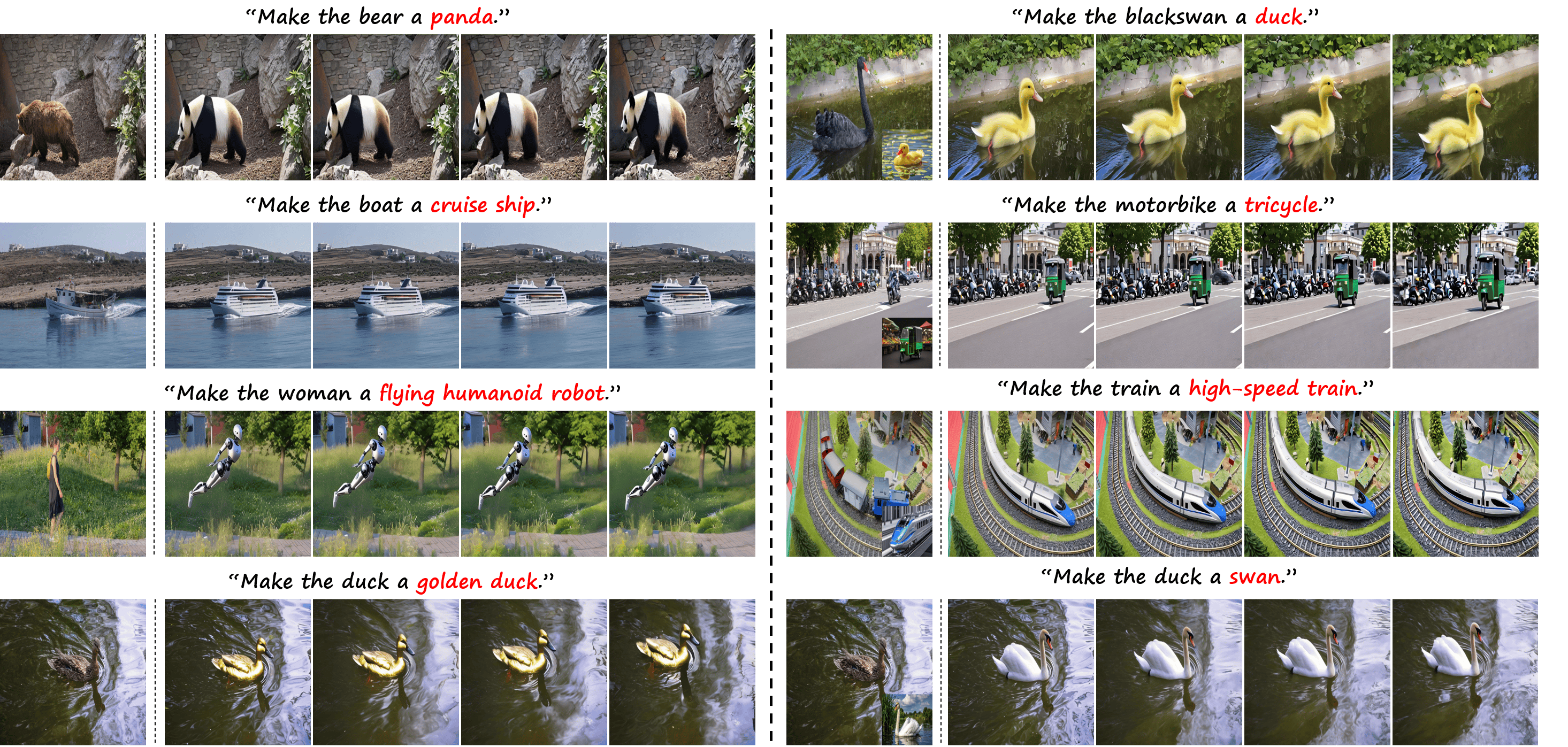}
  \vspace{-1.5em}
  \caption{
      More results under the text-based (left) and image-based (right) editing scenarios.
  }
  % \vspace{-0.8em}
  \label{fig: image-text-based-editing}
\end{figure}

\vspace{-0.3em}
In this section, we present additional qualitative results produced by FlowV2V, including face editing (Fig. \ref{fig: face-editing}), style transfer (Fig. \ref{fig: style-transfer}), rotation (Fig. \ref{fig: rotation}), multi-object scenarios (Fig. \ref{fig: multi-object}), instruction-based editing (Fig. \ref{fig: instruction-based}), text- and image-based editing (Fig. \ref{fig: image-text-based-editing}).

\vspace{-0.2em}
\subsection*{D. Impact of RAFT's Estimation Error}
\label{sec:raft}
\vspace{-0.3em}

In this section, we provide a detailed analysis of how the estimation errors of RAFT affect the performance of the FlowV2V model.
\vspace{-0.7em}
\paragraph{Textureless Regions.}
RAFT struggles in textureless regions where reliable motion cues are scarce. For example, when pale blue, nearly transparent tears slide down a face, the transparent liquid introduces almost no discernible texture and causes only minimal photometric variation against the background. As a result, RAFT fails to accurately estimate the optical flow in these regions.
%
%%%%%%%%%%%%%%%%%%%%%%%%%%%%%%%%%%%%%%%%%%%%%%

%%%%%%%%%%%%%%%%%%%%%%%%%%%%%%%%%%%%%%%%%%%%%%%
%%%%%%%%%%%%%%%%%%%%%%%%%%%%%%%%%%%%%%%%%%%%%%

%%%%%%%%%%%%%%%%%%%%%%%%%%%%%%%%%%%%%%%%%%%%%%%

%%%%%%%%%%%%%%%%%%%%%%%%%%%%%%%%%%%%%%%%%%%%%%%

%%%%%%%%%%%%%%%%%%%%%%%%%%%%%%%%%%%%%%%%%%%%%%%

%%%%%%%%%%%%%%%%%%%%%%%%%%%%%%%%%%%%%%%%%%%%%%%

\vspace{-0.7em}
\paragraph{Object Occlusion.}
RAFT encounters difficulties when objects become occluded or disappear between consecutive frames, as no subsequent visual information is available to support motion estimation. A typical case can be observed in the soccer ball example from the DAVIS-EDIT dataset, where the ball is occluded by trees, leading to inaccurate flow predictions and noticeable discrepancies between the generated and original videos.
\vspace{-0.7em}
\paragraph{Lighting Variation.}
RAFT is sensitive to significant variations in brightness or color between adjacent frames, which can substantially degrade feature matching accuracy. In particular, in face editing applications, lighting changes that occur as the subject’s mouth closes often lead to unreliable optical flow estimation, resulting in noticeable temporal inconsistencies and reduced visual quality in the generated video.
\vspace{-0.2em}
\subsection*{E. Efficiency Study}
\label{sec:efficiency}
\vspace{-0.3em}
In this section, we evaluate the average inference time of FlowV2V on the DAVIS-EDIT-C and DAVIS-EDIT-S datasets. The GPU used for these experiments was an H20-NVLink (96GB). As shown in Table \ref{tab:example}, FlowV2V achieves efficient inference at a resolution of 512×512, completing a video editing example in only about 2.5 minutes, markedly outperforming other methods in terms of speed. However, when the resolution is increased to 854×480, corresponding to the original video size, the average inference time nearly doubles. This increase is primarily due to the higher computational cost associated with the larger image resolution.

\vspace{-0.2em}
\subsection*{F. Limitations}
\label{sec:limitations}
\vspace{-0.3em}
Despite the strong performance and versatile applications demonstrated by FlowV2V, its effectiveness in some editing cases remains limited, primarily due to inherent restrictions imposed by the optical flow estimation module and the generative capability of the employed I2V backbone.

\end{document}